\documentclass[journal]{IEEEtran}
\usepackage[T1]{fontenc}% optional T1 font encoding
\usepackage{graphicx}
\usepackage{times}
\usepackage{helvet}
\usepackage{courier}
\usepackage{amsmath}
\usepackage{algorithm}
\usepackage{algorithmic}
\usepackage{csquotes} 
\usepackage{color}
\usepackage{paralist}
\usepackage{amssymb}
\usepackage{indentfirst}
\usepackage{subfigure}
\usepackage{float}
\usepackage{multirow}
\usepackage{cite}
\usepackage{mathrsfs}

\usepackage{mathrsfs}
\usepackage[colorlinks, linkcolor=red,  anchorcolor=blue, citecolor=blue]{hyperref}
\usepackage{textcomp,booktabs}
\usepackage{amssymb}% http://ctan.org/pkg/amssymb
\usepackage{pifont}% http://ctan.org/pkg/pifont

\usepackage{color}
\begin{document}

\title{Dynamic Attention guided Multi-Trajectory Analysis for Single Object Tracking}

\author{Xiao Wang, Zhe Chen, Jin Tang, Bin Luo, Yaowei Wang, \emph{Member IEEE}, Yonghong Tian, \emph{Senior Member IEEE}, Feng Wu, \emph{Fellow IEEE} 
\thanks{ Xiao Wang, Jin Tang and Bin Luo are with the Cognitive Computing Research Center of Anhui University, School of Computer Science and Technology, Anhui University, Hefei 230601, China. The main part of this work was done when the first author have a visit at The University of Sydney in 2019.  He is now a postdoc at Peng Cheng Laboratory, Shenzhen, China. 

Zhe Chen is with The University of Sydney, Australia. 

Yaowei Wang and Yonghong Tian are with Peng Cheng Laboratory, Shenzhen, China. They also with the National Engineering Laboratory for Video Technology, School of Electronics Engineering and Computer Science, Peking University, Beijing, China. 

Feng Wu is from University of Science and Technology of China, Hefei, China. 

Jin Tang and Zhe Chen are the first and second corresponding author, respectively. Email: \{tangjin, luobin\}@ahu.edu.cn, zhe.chen1@sydney.edu.au, \{wangx03, tianyh, wangyw\}@pcl.ac.cn, fengwu@ustc.edu.cn.}}

\markboth{IEEE Transactions on Circuits and Systems for Video Technology} 
{Shell \MakeLowercase{\textit{et al.}}: Bare Demo of IEEEtran.cls for IEEE Journals}

% make the title area
\maketitle

% As a general rule, do not put math, special symbols or citations in the abstract or keywords.
\begin{abstract}
Most of the existing single object trackers track the target in a unitary local search window, making them particularly vulnerable to challenging factors such as heavy occlusions and out-of-view movements. Despite the attempts to further incorporate global search, prevailing mechanisms that cooperate local and global search are relatively static, thus are still sub-optimal for improving tracking performance. By further studying the local and global search results, we raise a question: can we allow more dynamics for cooperating both results? In this paper, we propose to introduce more dynamics by devising a dynamic attention-guided multi-trajectory tracking strategy. In particular, we construct dynamic appearance model that contains multiple target templates, each of which provides its own attention for locating the target in the new frame. Guided by different attention, we maintain diversified tracking results for the target to build multi-trajectory tracking history, allowing more candidates to represent the true target trajectory. After spanning the whole sequence, we introduce a multi-trajectory selection network to find the best trajectory that deliver improved tracking performance. Extensive experimental results show that our proposed tracking strategy achieves compelling performance on various large-scale tracking benchmarks. The project page of this paper can be found at \textcolor{magenta}{\url{https://sites.google.com/view/mt-track/}}. 

\end{abstract}


\IEEEpeerreviewmaketitle

\section{Introduction}

\begin{figure}[!htb]
\center
\includegraphics[width=3.3in]{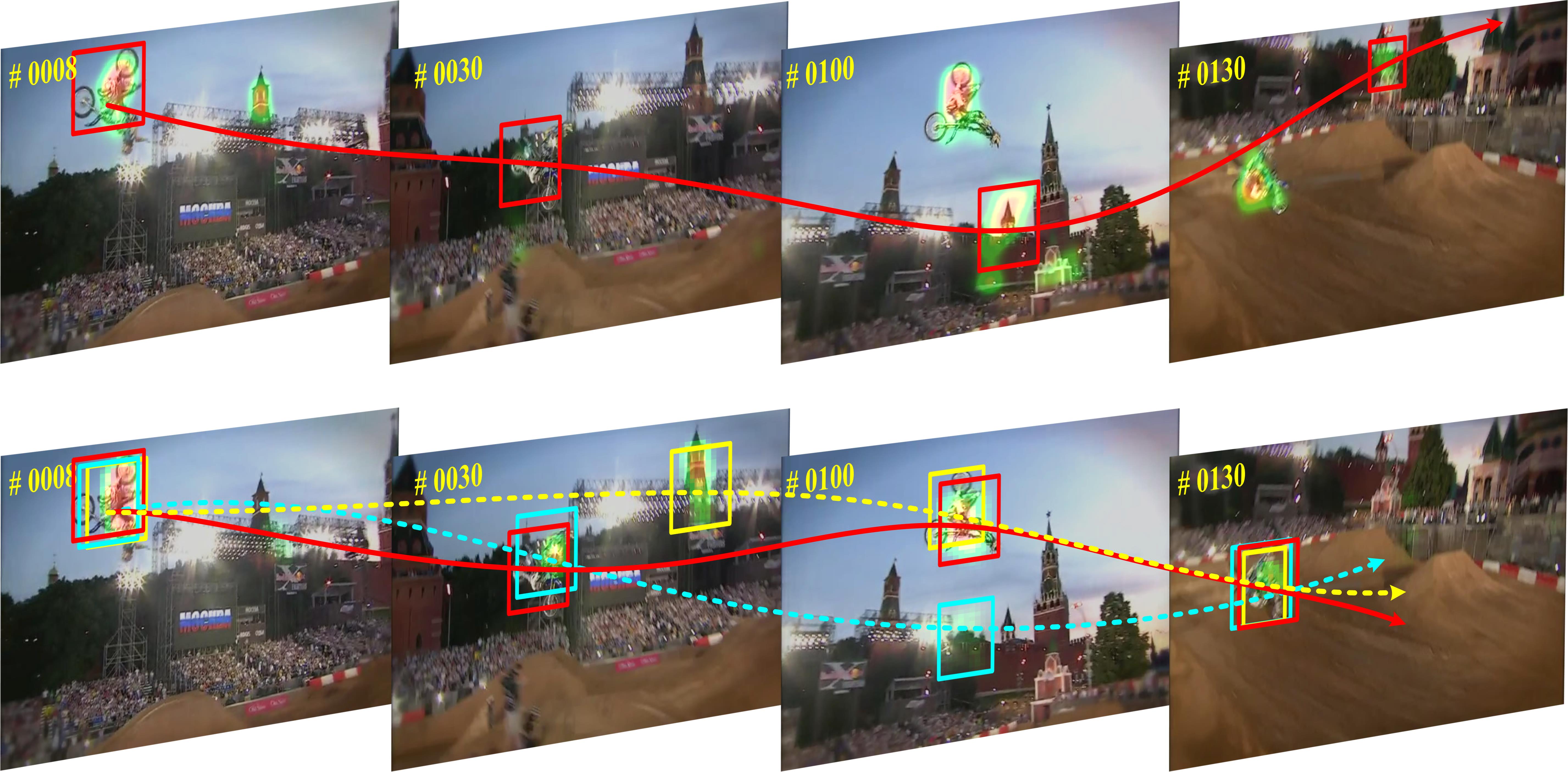}
\caption{ \textbf{The comparison between single trajectory tracking (the first row) and our dynamic target-aware attention guided multi-trajectory tracking framework (the bottom row).} Existing trackers usually estimate one location only for each frame, which may prone to model drift in challenging scenarios even the global search scheme adopted. Our proposed multi-trajectory tracking framework maintain multiple locations guided by dynamic target-aware attention for each frame and significantly increase the dynamics of baseline trackers. }
\label{motivation}
\end{figure}

\IEEEPARstart{W}{ith} the powerful representation ability of deep neural networks, existing single object visual trackers promisingly promote state-of-the-art tracking performance one after another under tracking-by-detection framework, benefiting many other practical applications, including UAV, robotics, video surveillance, and so on. However, even using cutting-edge deep neural networks \cite{zhipeng2019deeper} \cite{li2018siamrpn++}, current tracking algorithms are still poor under some challenging scenarios, such as heavy occlusions, out-of-view movements, fast motions, scale variations. This suggests that only improving deep visual features may reach a bottleneck for boosting tracking accuracy, which drives us to devise a novel and more advanced tracking strategy to tackle these challenging factors.

According to our observation, many of the existing single object trackers that follow the \emph{tracking-by-detection framework} generally attempt to track the target object, which is initialized in the first frame, based on  a unitary local search window. However, without modeling global information appropriately, such a local search strategy can be sensitive to the aforementioned challenging factors, thus visual trackers can be prone to drifting away. To handle these issues, some researchers propose to incorporate global search strategy into visual tracking \cite{yan2019skimming} \cite{wang2018describe} \cite{wang2019GANTrack} \cite{yang2019learning} \cite{zhu2016beyond}. Despite that current global search approaches can handle the issues of the local search tracking strategy to some extents, they still suffer from the dilemma of selection between local and global search for each frame. Moreover, it is also difficult for current approaches that introduce global information to accurately locate the potential regions that contain the target, especially when multiple similar objects appeared in the image simultaneously.

With deeper investigation, we find that current approaches, even those that cooperate local and global search results, only model target appearance and estimate target status in a relatively static way, i.e., the tracking paradigms for each frame is fixed and static to a single bounding box. In other words, current visual trackers can be easily distracted by similar objects or noisy backgrounds when challenging factors like occlusions and out-of-view movements occur in the sequence if the target appearance modeled by the tracker is corrupted in the estimated bounding box. Furthermore, in existing trackers, the noise of an unreliable tracking result at a frame could also be continuously accumulated in later frames when the estimated target bounding box at each frame drift away from true position. This easily leads to more severe drifting.
Even though some studies \cite{ma2015long} \cite{zhang2018learning} \cite{hong2015multi} introduce re-detect mechanism to correct the inaccurate tracking results once the target object is found to be lost, the already failed tracking results before re-detection will not be corrected anyway, and these erroneous tracking results still degrade the overall tracking performance. To tackle these problems, we instead seek to explore a novel search strategy that can allow more dynamics for single object tracking.

We are inspired by the recent progress of image captioning \cite{hossain2019comprehensive} \cite{bai2018survey} task, in which given images are generally embedded into deep representations, and a sequential model is applied to deliver corresponding language descriptions based on the embedded representations. In particular, by maintaining multiple words at each time step, the decoding stage provides more adequate language descriptions by applying beam search algorithm to allow more dynamics when analyzing the maintained words. We believe that the visual trackers can also take advantage of such a strategy that searches better results through different candidates, if multiple different tracking states and trajectories are maintained during tracking.

\begin{figure}[!htb] 
\center 
\includegraphics[width=3.5in]{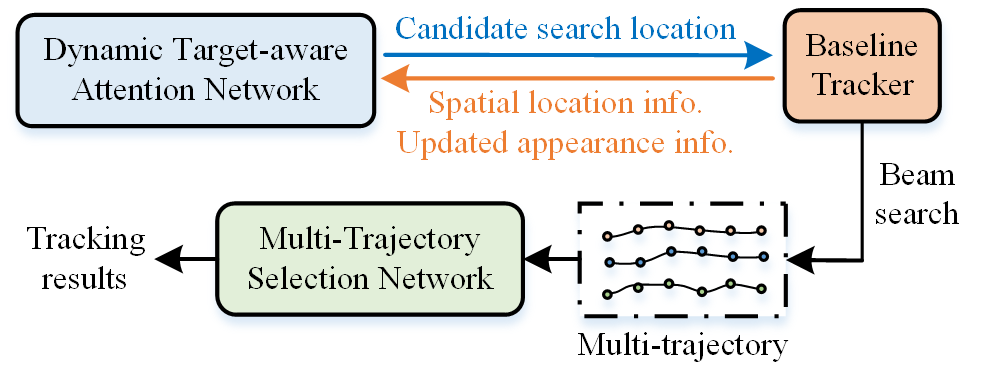}
\caption{ \textbf{The pipeline of our proposed dynamic target-aware attention guided multi-trajectory tracking.} } 
\label{pipeline}
\end{figure}

In this study, we propose a novel dynamic attention-guided multi-trajectory tracking framework to introduce more dynamics for tackling various challenging tracking issues. As shown in Fig. \ref{pipeline}, our tracker consists of three main modules, i.e., a novel dynamic target-aware attention network to guide the global search for the target, a baseline tracker for local search about the target, and a multi-trajectory selection network. 
%Given a video sequence, we first conduct tracking in a local search window with Siamese tracker (THOR \cite{sauer2019tracking} used in this paper). Meanwhile, we apply our proposed dynamic target-aware attention network to achieve more effective global search, benefiting long-term tracking performance. 
Different from existing works \cite{wang2018describe} \cite{wang2019GANTrack} \cite{yang2019learning}, our dynamic target-aware attention network maintains multiple target templates and fully utilizes hierarchical semantic feature representations obtained from different convolutional layers to provide more robust appearance model for the target and more accurate candidate regions when performing global search. Meanwhile, in each frame, the baseline tracker is employed to conduct local search for the target.
By performing a joint local and global search for visual tracking in a parallel manner for each frame, we obtain and maintain multiple candidate states for the target. After spanning the whole video sequence, we employ a multi-trajectory selection network to find the best tracking trajectory based on different maintained tracking results for the sequence. 

To sum up, the contributions of this paper can be summarized in the following three aspects: 

$-$ We propose a novel dynamic attention-guided multi-trajectory based tracking framework for single object tracking. In particular, based on multiple tracking trajectories maintained for the target, our proposed approach introduces more dynamics for cooperating local and global tracking results to help tackle the challenging tracking issues effectively. 

$-$ We propose a dynamic target-aware attention network to build a more robust and dynamic appearance model for global search based tracking. We also devise a novel multi-trajectory selection network to estimate the best tracking results according to maintained multi-trajectory information. 

$-$ Our proposed tracking strategy achieves compelling performance on several large-scale tracking benchmarks, validating the effectiveness of our proposed method.

\section{Related Work} 

\textbf{Siamese Network based Trackers.}
Many trackers are developed based on Siamese network due to its high accuracy and efficiency. Tao \emph{et al.} \cite{Tao2016Siamese} propose to use the Siamese network to learn a fixed matching function and tracking the target object without any update. Wang \emph{et al.} \cite{Wang_2018_CVPR} further developed this framework by introducing adversarial samples for robust visual tracking. Bertinetto \emph{et al.} \cite{Bertinetto2016SiameseFC} use a fully convolutional Siamese network for tracking by measuring the region-wise feature similarity between the target object and candidates (named SiamFC). GOTURN is proposed by Held \emph{et al.} \cite{held2016GOTURN} which uses a motion prediction model developed based on Siamese network for high-speed tracking. Recently, many researchers focus on designing a more powerful network for Siamese based tracker, such as SiamDW \cite{zhipeng2019deeper}, SiamRPN++ \cite{li2018siamrpn++}. Some works improved the Siamese network based tracker by cascaded region proposal networks \cite{fan2019siamese}, relation reasoning \cite{gao2019graph}, dynamic target template update \cite{Yang_2018_ECCV} \cite{sauer2019tracking}, re-detection \cite{voigtlaender2020siam}, meta-learning \cite{wang2020tracking, choi2019deepmeta}, and others \cite{chen2017generic}.  
Although these works all improved the Siamese network based tracker from different perspectives, however, they still adopt the greedy search based strategy for visual tracking. Therefore, these trackers may be sensitive to the challenging factors we mentioned above. In this paper, we propose a novel multi-trajectory tracking framework for visual tracking which can address issues caused by greedy search to some extent.

\textbf{Long-term Tracking.} 
Existing long-term tracking algorithms usually introduce \emph{re-detection} mechanism in the tracking procedure. Kalal \emph{et al.} \cite{kalal2011tracking} propose a tracking-learning-detection framework for long-term tracking. They utilize optical-flow based matcher for local search and also adopt the ensemble of weak classifiers for re-detection. Ma \emph{et al.} \cite{ma2015long} propose a long-term correlation filter which use KCF for local tracking and external random ferns classifier for long-term detector. Valmadre \emph{et al.} \cite{valmadre2018long} develop a long-term tracker named SiamFC+R, which also integrates a simple re-detection scheme with SiamFC and conduct re-detection when the SiamFC's response is lower than a pre-defined threshold. Some works attempt to address the tracking task with \emph{key point matching} \cite{nebehay2015clustering} \cite{hong2015multi} or \emph{global proposal mechanism} \cite{zhu2016beyond} \cite{wang2018describe} \cite{wang2019GANTrack} \cite{yang2019learning}. However, as noted in \cite{yan2019skimming}, the keypoint extractors and descriptors are not reliable in complex environments, which may limit their overall performance. Zhu \emph{et al.} \cite{zhu2016beyond} propose the EBT tracking based on EdgeBox \cite{zitnick2014edge} and improve the baseline method significantly. Yan \emph{et al.} \cite{yan2019skimming} propose a `Skimming-Perusal' tracking framework for real-time and long-term tracking. They use a switching mechanism to decide local or global proposals should be used in each frame and achieve better results on OxUvA dataset.   Fan \emph{et al.} \cite{fan2017parallel} propose a parallel tracking and verifying framework (PTAV) which can achieve better results on UAV20L dataset. Wang \emph{et al.} \cite{wang2018describe} \cite{wang2019GANTrack} \cite{yang2019learning} develop the target-aware attention mechanism for global proposal generation and integrate with MDNet for robust tracking. GlobalTrack is proposed in \cite{huang2019globaltrack} which only employ global search on whole video frames also achieves good performance on long-term datasets. Dai \emph{et al.} \cite{dai2020LTMU} propose offline-trained meta-updater to address the update issue in long-term tracking.

\textbf{Trajectory-based Tracking.} 
Although very few, there are still some related works that exploit trajectory information to achieve tracking. For example, MTA \cite{lee2015multihypothesis} is proposed to conduct tracking by trajectory selection using tracking results obtained from STRUCK \cite{hare2016struck}. In practice, the MTA is subject to poor hand-crafted features and empirical multi-trajectory analysis that can not be optimized. In addition to MTA, researchers also develop MHT \cite{kim2015MHT} to conduct multi-object tracking based on different trajectories. To track multiple objects, the MHT designs a track tree construction and updating scheme to allow more dynamics. However, there are many differences between our proposed method and MHT. First, MHT relies on hand-crafted features and requires detected bounding boxes, while our framework is built upon more powerful deep convolutional features and does not require the detection model to provide bounding box results. Second, MHT uses tree structure to derive trajectories, while our method maintain independent tracking results and derive trajectories based on beam search strategy which can allow more trajectories to describe a single target. Also, MHT relies on detection models to encode target appearances, while our method can dynamically encode appearances. Lastly, rather than MHT that performs empirical tree construction algorithm to deliver trajectory results, our proposed multi-trajectory selection network can be optimized to select the best trajectory result.

\section{Our Proposed Approach} \label{MT_Algorithm}

In this section, we will first give an introduction to the baseline tracker THOR used in this work. Then, we will give an overview of our proposed modules. After that, we will dive into the details on dynamic attention model, trajectory selection network. We mainly focus on the motivation, detailed network architecture, and advantages of our proposed modules for object tracking.

\subsection{Preliminary: THOR}
THOR is a Siamese network based visual tracker proposed by Axel Sauer \emph{et al.} \cite{sauer2019tracking}. They design a simple but effective target template update mechanism for tracking procedure. Specifically, they utilize two modules, i.e., the short-term module (STM) and long-term module (LTM) to store the tracked results.  The authors define a novel diversity measure in the space of Siamese features to select the most diverse templates. 

For the long-term module, they try to maximize the volume $\Gamma (f_1, ... , f_n)$ of the parallelotope formed by the feature vectors $f_i$ of the template $T_i$. They use convolutional operation to compute the similarity between different templates in memory and obtain a Gram matrix: 
\begin{equation}
\label{GramMatrix}
G(f_1, ... , f_n) = \\ 
\begin{bmatrix}
 &f_1 \star f_1  &f_1 \star f_2  & ... & f_1 \star f_n\\ 
 &\vdots   &\vdots  &\ddots       &\vdots \\ 
 &f_n \star f_1  &f_n \star f_2  & ... & f_n \star f_n 
\end{bmatrix}
\end{equation}
where $G$ is a square $n \times n$ matrix. Therefore, the Gram determinant (i.e., the determinant of $G$) can be written as: 
\begin{equation}
\label{GramDeterminant} 
\max_{f_1, ... , f_n} \Gamma (f_1, ... , f_n) \varpropto  \max_{f_1, ... , f_n} |G(f_1, ... , f_n)| 
\end{equation}
The template can be incorporated into the memory, if the Gram determinant can be increased when replacing one of the allocated templates.

The short-term module is introduced to handle abrupt movements and partial occlusion. They update the memory slots of STM in a first-in, first-out manner. Different from LTM, they compute the diversity measure $\gamma$ of STM as follows: 
\begin{equation}
\label{stmFunction} 
\gamma = 1 - \frac{2}{N(N+1) G_{st, max}}  \sum_{i < j}^{N} G_{st, ij}
\end{equation}
The authors integrate the LTM and STM module with three Siamese network based trackers (SiamFC \cite{bertinetto2016fully}, SiamRPN \cite{li2018siamRPN} and SiamMask \cite{wang2019fast}), and achieve better tracking performance than the baseline methods. More details of THOR can be found in \cite{sauer2019tracking}.

THOR works well on short-term and small datasets, the performance on large-scale and long-term tracking datasets are still unknown. This tracker also adopts a greedy search in a local search window which makes their performance unsatisfied on challenging benchmarks. In this paper, we introduce a novel dynamic target-aware attention mechanism and integrate with SiamRPN based THOR for robust tracking in the multi-trajectory manner.

\subsection{Overview}
In general, our proposed method consists of three key components, including \emph{the dynamic target-aware attention network}, \emph{a local search-based baseline tracker}, and \emph{the multi-trajectory selection network}. Given a testing video sequence, we first perform local tracking based on a Siamese tracker (e.g. THOR \cite{sauer2019tracking}) to search the target according to the response map of a local region. In the meantime, we perform global search to locate the target within the whole frame. To tackle the challenging factors, such as heavy occlusions, fast motions, and out-of-view movements, we introduce a novel dynamic target-aware attention network to help construct a more robust and dynamic appearance model, facilitating the global tracking to better locate the target. Moreover, rather than cooperating local and global search results in a relatively static way, we propose to maintain multiple different tracking results obtained from local and global search in a parallel manner to achieve multi-trajectory tracking with more dynamics. At the end of testing video, with the help of a carefully designed trajectory selection network, we select the trajectory with the highest confidence score to deliver a further refined final tracking results.

\subsection{Dynamic Target-Aware Attention Network}

\subsubsection{Our Approach}
\begin{figure}[!htb]
\center
\includegraphics[width=3.5in]{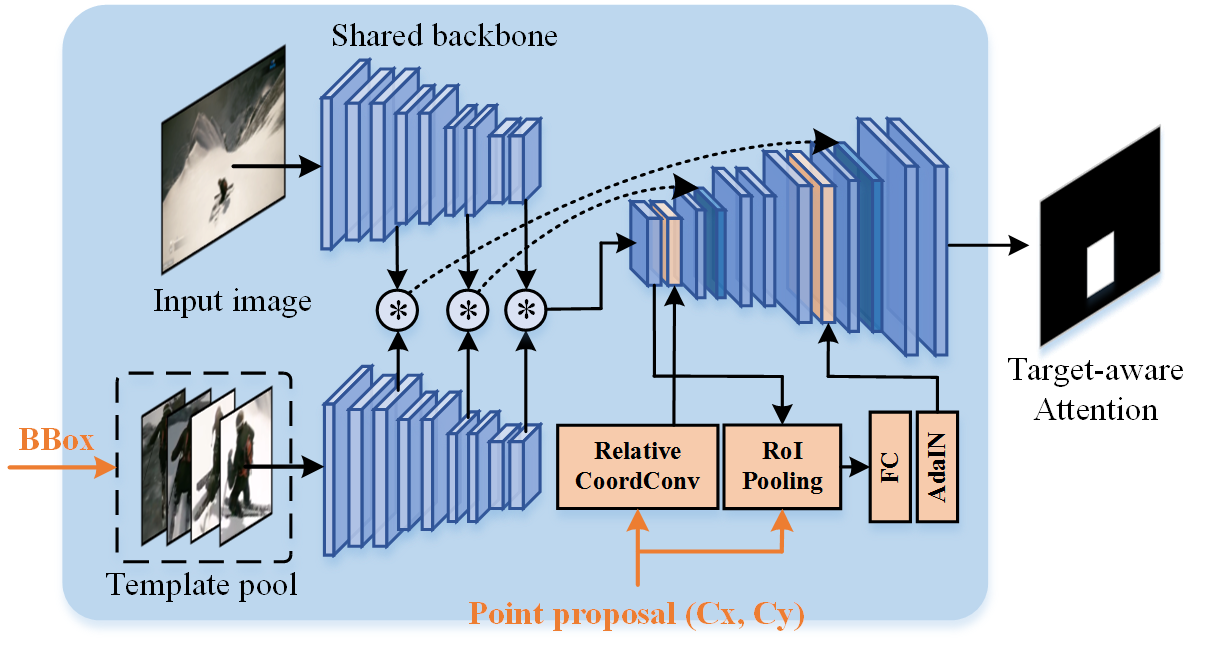}
\caption{ \textbf{Illustration of our dynamic target-aware attention generation module. }}
\label{pipeline_dyTANet}
\end{figure}

We proposed the dynamic target-aware attention network to construct dynamic appearance model by maintaining multiple target templates. In particular, we first extract feature maps from three different convolutional modules to deliver hierarchical and more powerful representations. Then, we treat target features as convolutional filters. We employ these filters to perform convolution on the feature maps of the whole image. The output features will be later fed into a gate layer which is used to control the information flow. Afterwards, the gated features will be gradually fed into a decoder network and output corresponding attention map. Inspired by \cite{sofiiuk2019adaptis}, we incorporate the \emph{point proposals} to make the attention map focus more on the target regions encoded by the coordinates of the bounding box corners to handle the issues caused by similar target objects. In our proposed method, we encode point proposals into horizontal and vertical feature maps with relative CoordConv \cite{sofiiuk2019adaptis} and regard it as prior information to estimate the current attention map. Also, we apply an RoI pooling layer to extract the point feature and feed it into an adaptive instance layer \cite{huang2017arbitrary} to achieve instance-level attention estimation. Moreover, we employ the local search-based tracker to help collect and maintain a set of short-term and long-term target templates (i.e., the dynamic template pool in Fig. \ref{pipeline_beamtracker}) to provide dynamic target feature representations during the tracking procedure. In this way, we can utilize the updated target templates to conduct dynamic target-aware attention estimation. In general, the proposed dynamic target-aware attention network can provide more accurate global search regions than the static counterpart methods \cite{wang2018describe} \cite{wang2019GANTrack} \cite{yang2019learning}. An overview of our attention network is given in Fig. \ref{pipeline_dyTANet}.

\subsubsection{Network Architecture} 
Given a video frame $I$ and a target template $T$, we first resize them into $300 \times 300$ and $100 \times 100$, respectively. Then, we put them into an encoder network with two branches. Each branch is a residual network (ResNet-18 used in this paper) which is pre-trained on ImageNet classification task. To obtain a more effective feature representation, we extract the hierarchical semantics from feature maps of both inputs from three convolutional blocks whose dimension are [$128\times 38\times 38$; $256\times 19\times 19$; $512\times 10\times 10$] and [$128 \times 13 \times 13; 256 \times 7 \times 7; 512 \times 4 \times 4$], respectively. We use $F_i^{T}$ and $F_i^{I}$ to denote the $i^{th}, i \in \{1, 2, 3\}$ feature map of target object and global image. Different from previous works which directly concatenate the two feature maps, we conduct convolutional operation with target feature $F_i^{T}$ on global image features $F_i^{I}$ to boost their interaction: 
\begin{equation}
\small 
\label{dcovOperation} 
F_i^{TI} = F_i^{T} \star  F_i^{I}
\end{equation}
where $\star$ denotes the convolutional operator. Then, $F_i^{TI}$ is fed into a gate layer to achieve controllable information flow which is widely used in recurrent neural network (such as forget gate, input gate in LSTM \cite{hochreiter1997lstm}). This process can be written as: 
\begin{equation}
\label{gateOperation} 
\small 
\hat{F_i^{TI}} =  F_i^{TI} \odot \sigma (F_i^{TI})
\end{equation}
where $\sigma$ is a sigmoid gate layer and the $\odot$ denote dot product between two matrix. These features will be fed into a decoder network gradually using skip connections to predict corresponding target-aware attention map.

Global search-based tracking that only utilizes static target templates could suffer from the issue of the existence of multiple similar target objects in the image. Since the global search module will provide multiple search regions for different similar target objects, it can be very difficult for the tracker to recognize the true target. In this paper, we propose to utilize \emph{point proposal} ($x, y$) to help target-aware attention estimation for a more robust global search. The center location of the previous tracking result is considered as the point proposal. Specifically, the point proposal is used in two ways: the \emph{location prior} of target object and \emph{``characteristic" embedding}.   For the \emph{location prior}, existing works \cite{novotny2018semi} \cite{sofiiuk2019adaptis} \cite{liu2018intriguing} all shown that the pixel coordinates are important to disambiguate different object instances. Liu \emph{et al.} \cite{liu2018intriguing} proposed a simple CoordConv layer to encode the pixel coordinates into feature maps by creating a tensor of the same size as input that contains pixel coordinates normalized to [-1, 1]. The authors of \cite{sofiiuk2019adaptis} further proposed a \emph{Relative CoordConv} block which can utilize standard backbone networks for their task, such as pre-trained ResNet. Following \cite{sofiiuk2019adaptis}, we also construct two coordinate feature maps, one for \emph{x}-coordinates and one for \emph{y}-coordinates. The values of constructed $x$-map vary from $-1$ to $+1$, where $-1$ and $+1$ corresponding to $x-R$ and $x+R$, respectively. The $y$-map has similar attributes and the $R$ is a hyper-parameter used to denote the radius of Relative CoordConv. The constructed prior map is concatenated with feature maps produced by the regular convolutional network. 
Besides, we also attain the ``\emph{characteristic}" embedding based on the point proposal. More specifically, given the point proposal ($x, y$), we extract a $1 \times 1 \times 512$ feature $Q(x, y)$ with RoI pooling on the feature maps generated from the third convolutional layer of decoder network. This feature vector is then fed into two fully connected layers and the AdaIN layer \cite{huang2017arbitrary} to achieve instance selection. The output feature will be integrated into the decoder network.

Using the template representation collected and maintained by the local search-based baseline tracker to achieve Siamese tracking, our attention estimation can be implemented in a \emph{dynamic} manner, while previous works are all \emph{static} target-aware attention. In other word, we can borrow the updated target representation in the tracking procedure for more accurate attention estimation. Adaptively switching between local and global search is one intuitive approach for robust tracking as many previous trackers do \cite{yan2019skimming, wang2019GANTrack, wang2018describe, yang2019learning}. To maximize the benefits of dynamic appearance model for the whole video sequence, our tracker runs in a batch manner.

\subsubsection{Advantages of Our Approach}
%Although local search-based tracking is already promising, it can be easily affected by noise and errors in a sequence. To this end, global search-based tracking is important to compliment local search-based tracking by re-locating the target. As described above, we mainly apply target-aware attention \cite{wang2019GANTrack} to achieve global search-based tracking. 
%%In practice, the target-aware attention can provide an attention map for the frame to focus on the target areas based on the target template initialized in the first frame and global video frame.

%By addressing the issues of the original target-aware attention methods \cite{wang2018describe} \cite{wang2019GANTrack} \cite{yang2019learning} in this study, it is crucial to construct a robust target-aware attention model for global search-based tracking, since local search-based tracking could be easily affected by noise and errors in a sequence and global search-based tracking is the key to re-locate the target. We mainly apply target-aware attention \cite{wang2019GANTrack} to achieve global search-based tracking. In practice,  the target-aware attention can provide an attention map for the frame to focus on the target areas based on the target template initialized in the first frame and global video frame.
%Despite descent performance, original target-aware attention-based tracking is still sub-optimal to fulfill global search-based tracking. We observe that the original method have the following issues. 

Compared with original target-aware attention based trackers, the highlights of our proposed dynamic attention model can be listed as follows:
\textbf{Firstly}, the original method exploits the features of the target object and global image based on simple concatenation operation without considering the relationships and interactions between the two types of features. This may limit the representational power of the model. 
\textbf{Secondly}, they only utilize the feature map from the last convolutional layer of their encoder, which can not fully utilize the hierarchical semantic information for better attention prediction. 
\textbf{Thirdly}, they only utilize the feature map of the target object initialized in the first frame. However, since the appearance of the target is generally continuously changing in a tracking sequence, such a fixed appearance model is almost infeasible to track the target that undergoes significant appearance variations in the corresponding video sequence. 
\textbf{Last but not the least}, their attention model is also too primitive to handle multiple similar target objects. In practice, there would be multiple high response regions which may easily distract the tracker on the similar but non-target objects.

As a result, the aforementioned issues inspire us to design an advanced new attention scheme for tracking in challenging scenarios, especially for long-term object tracking. In particular, we propose the dynamic target-aware attention mechanism, as shown in Fig. \ref{pipeline_dyTANet}, to introduce more dynamics in the global search-based tracking procedure. By further addressing the issues of the original target-aware attention methods \cite{wang2018describe} \cite{wang2019GANTrack} \cite{yang2019learning} in this study, our method constructs a more robust and more dynamic target-aware attention model for global search-based tracking.

\begin{figure}[!htb]
\center
\includegraphics[width=3.5in]{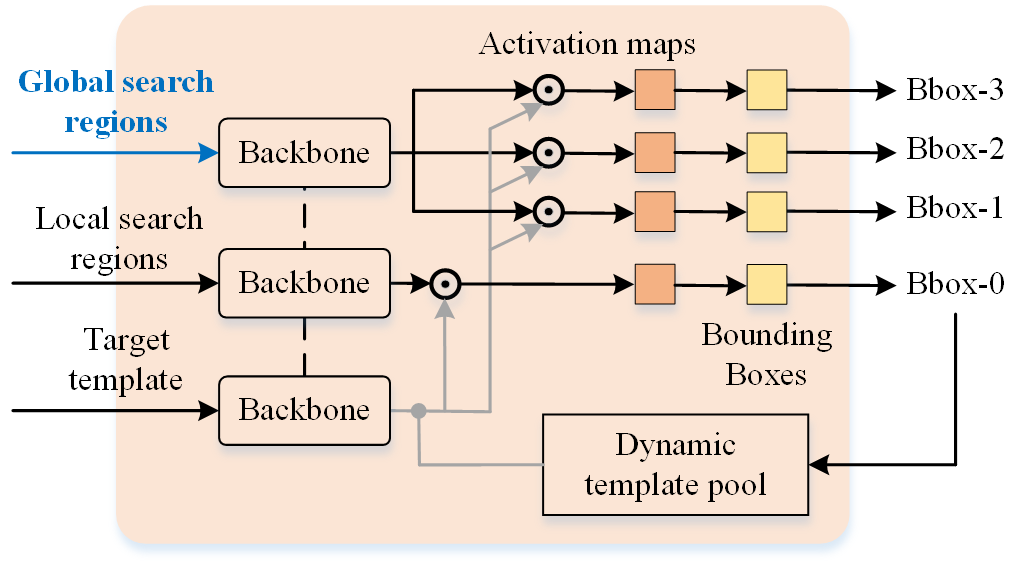}
\caption{ \textbf{Illustration of our attention guided multi-trajectory tracking module}. }
\label{pipeline_beamtracker}
\end{figure}

\subsection{Multi-Trajectory Selection Network}
Based on aforementioned dynamic attention model, we explore the beam search strategy to replace greedy search policy for visual tracking. Specifically, we keep multiple different tracking results during tracking and find the best trajectory to approach robust tracking. In particular, in each frame, instead of simply selecting the location/proposal with the highest confidence score as the final tracking result, we keep record of multiple candidate tracking results in a \emph{beam search} manner: 
% 
% first, we obtain the candidate regions from the global search-based tracking by utilizing the target templates collected from the local search-based tracking; 
first, we obtain the candidate regions from the global search module, i.e., the dynamic target-aware attention network.
Then, we locate the target from these regions with the local search-based tracker and select the most reliable top-$k$ results as the current search results, obtaining $k$ potential trajectories, as shown in Fig. \ref{pipeline_beamtracker}. Similar operations are executed for the subsequent video frames until the end of the video. After this procedure is completed, we measure the quality of tracking results in each frame with the trajectory selection network. A new trajectory with a maximum selection score will be chosen as the final tracking result of the current test video.

\begin{figure}[!htb]
\center
\includegraphics[width=3.5in]{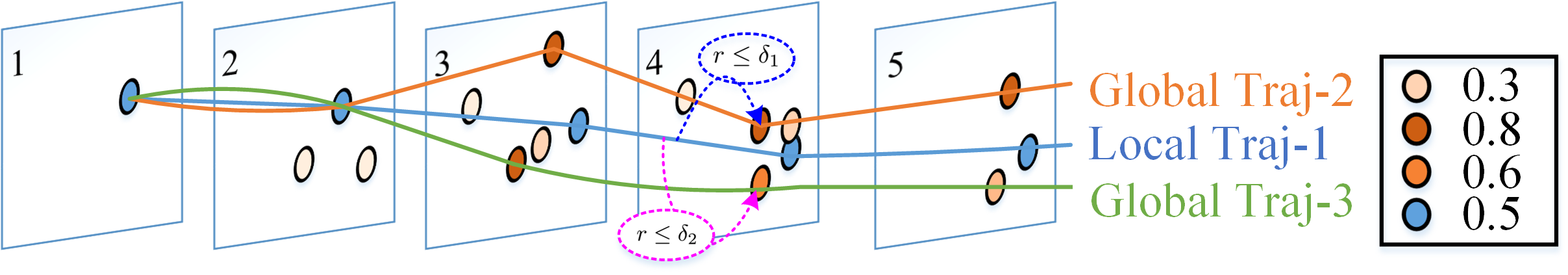}
\caption{ \textbf{Illustration of tracking by multi-trajectory selection. Best viewed by zooming in}.}
\label{beamSearch}
\end{figure}

For example, suppose that a video has T frames and each frame has 3 candidates, we always maintain 3 trajectories for tracking, since introducing more candidates will cost excessively large computational time. As illustrated in Fig. \ref{beamSearch}, two threshold parameters $\delta_1$ and $\delta_2$ are used to measure the quality of current tracking results for 3 trajectories. We denote Traj-1 as the local search result. We also denote $r$ as the confidence score of the local search based tracker. For each frame, if $r \leq \delta_1$, the search region will be switched into global attention region with best similarity (i.e., the dark orange points in Fig. \ref{beamSearch}, different colors means various confidence), therefore we have a new trajectory Traj-2. Meanwhile, if $r \leq \delta_2$, we search the target object from another global attention region if existed, therefore, we can attain the Traj-3. The Traj-1, 2, 3 then form the 3-trajectory search. This procedure shares a similar idea with beam search which is a heuristic search algorithm that explores a graph by expanding the most promising node in a limited set and widely used in image caption \cite{aneja2018convolutional}. Our proposed multi-trajectory selection policy and beam search all attempt to maintain multiple candidates for final selection. It is easy to find that our tracker can conduct joint local and global search in an adaptive manner for robust tracking and our model can still work well on long-term videos.

\begin{figure}[!htb]
\center
\includegraphics[width=3.3in]{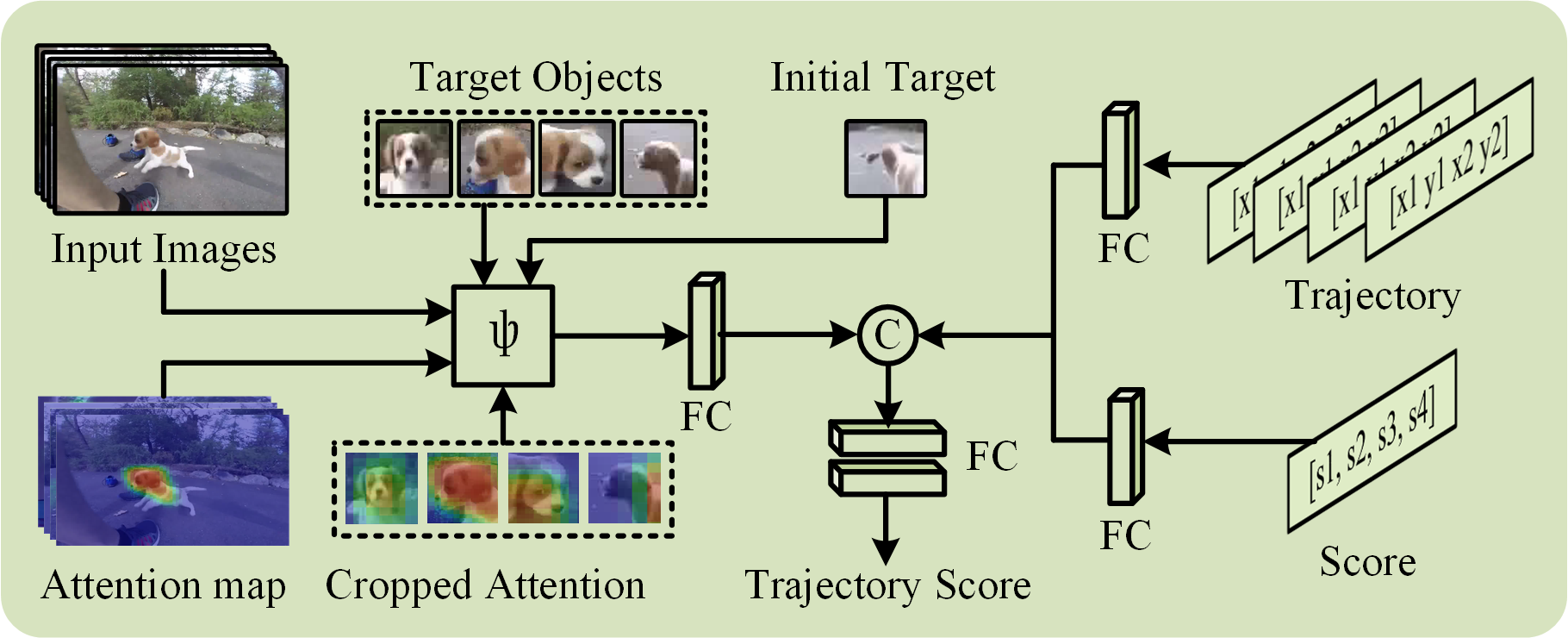}
\caption{ \textbf{Our proposed trajectory selection module}. }
\label{criticNet}
\end{figure}

After performing multi-trajectory based tracking through the whole video, as shown in Fig. \ref{criticNet}, we take all the available information as the input of the multi-trajectory selection network to accurately predict the quality of each trajectory. The encoded information includes input video frames, initial target template, predicted attention map, the cropped tracking result and corresponding target-aware attention patch, the specific value of predicted bounding box, and its similarity score predicted by the Siamese tracker. For all the input patches, we resize them into $300 \times 300$ and then feed them into a residual network. By aggregating these features via concatenation and reshaping, we send the aggregated features into a fully connected layer for dimension reduction. For bounding boxes and scores, we utilize two fully connected layers to encode them into corresponding feature vectors, respectively. Then, these feature vectors will be concatenated and fed into two fully connected layers for regression. The ground truth for this regression task is the IoU (Intersection over Union) between the predicted trajectory and ground truth annotations. In our implementation, we adopt GIoU \cite{rezatofighi2019generalized} which is an improved evaluation metric between two bounding boxes.

\section{Experiments}

\subsection{Dataset and Evaluation Metric}
In this paper, we train our dynamic target-aware attention network on two tracking datasets: TLP \cite{moudgil2018long} and DTB \cite{li2017visualuav}, which totally contain 120 video sequences. The trajectory selection network is trained on the training subset of GOT-10k dataset \cite{huang2018got}. We test our model on several popular tracking benchmarks, including OTB-2015 \cite{wu2015object}, GOT-10k \cite{huang2018got}, OxUvA \cite{valmadre2018long}, LaSOT \cite{fan2019lasot}, VOT2018-LT \cite{kristanVOT2018lt}, UAV123 \cite{mueller2016benchmarkuav20l} and UAV20L \cite{mueller2016benchmarkuav20l}. A brief introduction to these benchmark datasets are given below.

\textbf{OTB-2015 \cite{wu2015object}} contains 100 video sequences and also defines 9 attributes such as \emph{Illumination Variation, Scale Variation, Occlusion, Deformation}. It is one of the most widely used benchmark datasets for visual tracking since its release in 2015.

\textbf{GOT-10k \cite{huang2018got}} is constructed based on the backbone of WordNet structure \cite{miller1995wordnet}. It populates the majority of over 560 classes of moving objects and 87 motion patterns. It contains 10,000 videos totally, with more than 1.5 million manually labeled bounding boxes. The authors select 280 videos as the test subset and the rest of videos are used for training.

\textbf{OxUvA \cite{valmadre2018long}} is developed for the training and evaluation of long-term trackers. It comprises 366 sequences spanning 14 hours of video which can be categorized into 22 classes. This dataset is divided into train and testing subset, which contains 200 and 166 videos respectively.

\textbf{LaSOT \cite{fan2019lasot}} is the currently the largest long-term tracking dataset which contains 1400 video sequences with more than 3.5M frames in total. The average video length is more than 2,500 frames and each video contains challenging factors deriving from the wild, e.g., out-of-view, scale variation. It provides both natural language and bounding box annotations which can be used for the explorations of integrating visual and natural language features for robust tracking. For the evaluation of LaSOT dataset, \emph{Protocol I} employs all 1400 sequences for evaluation and \emph{Protocol  II} uses the testing subset of 280 videos.

\textbf{VOT2018-LT \cite{kristanVOT2018lt}} is a long-term dataset which contains 35 videos with a total length of 146817 frames. It is calculated that the target object will disappear for 12 times and each lasting on average 40 frames for each video.

\textbf{UAV123 \cite{mueller2016benchmarkuav20l}} and \textbf{UAV20L \cite{mueller2016benchmarkuav20l}} is an aerial video dataset which designed for low altitude UAV target tracking. It is consisted of 123 videos comprising more than 110K frames. They also provide a high-fidelity real-time visual tracking simulator for evaluation. The authors also merge these subsequences and pick the 20 longest sequences for long-term evaluation, also termed UAV20L.

For OTB-2015 \cite{wu2015object}, GOT-10k \cite{huang2018got}, LaSOT \cite{fan2019lasot}, UAV123 \cite{mueller2016benchmarkuav20l} and UAV20L \cite{mueller2016benchmarkuav20l}, \textbf{Precision Plots} and \textbf{Success Plots} are adopted for the evaluation (also termed \textbf{PR} and \textbf{SR}). The first evaluation metric illustrates the percentage of frames where the center location error between the object location and ground truth is smaller than a pre-defined threshold (20-pixel threshold are usually adopted). The second one demonstrates the percentage of frames the Intersection over Union (IoU) of the predicted and the ground truth bounding boxes is higher than a given ratio. It is worthy to note that the \textbf{AO} is also adopted for the evaluation of GOT-10k \cite{huang2018got} dataset. The AO denotes the average of overlaps between all ground truth and estimated bounding boxes. The VOT2018-LT \cite{kristanVOT2018lt} dataset adopts \textbf{Precision, Recall} and \textbf{F1-score} for the evaluation. Specifically, the definition of these metrics are: 
\begin{equation}
\label{Precision}
\small  
Precision = \frac{TP}{TP + FP}, ~~~~Recall = \frac{TP}{TP + FN}
\end{equation}
\begin{equation}
\label{Precision}
\small  
F1-score =  \frac{2 \times Precision \times Recall}{Precision + Recall}
\end{equation}
where TP, FP and FN are used to denote the True Positive, False Positive and False Negative, respectively. 
For the OxUvA \cite{valmadre2018long} long-term tracking benchmark, the \textbf{TPR, TNR} and \textbf{MaxGM} are adopted for the evaluation. Specifically, the TPR gives the fraction of present objects that are reported \emph{present} and correctly located, while TNR gives the fraction of absent objects that are reported \emph{absent}. The MaxGM is a single measure of tracking performance which can be formulated as: 
\begin{equation}
\small 
\label{maxGMFunction}
MaxGM = \mathop{\max}_{0 \leq p \leq 1} \sqrt{((1-p) * TPR) ((1-p) * TNR + p)}. 
\end{equation}

\subsection{Implementation Details}
In this paper, we utilize binary cross-entropy loss to train the dynamic target-aware attention network. The ground truth mask is obtained from existing tracking datasets by whiting the target object and black the background regions as previous works do \cite{wang2018describe} \cite{wang2019GANTrack} \cite{yang2019learning}. The batch size is $20$, learning rate is $1e-4$. For the training of the trajectory selection network, we first run our target-aware attention guided THOR to collect the predicted trajectories and attention maps on selected $3k$ video sequences from the training subset of the GOT-10k dataset. After that, we train this network for 50 epochs on the collected dataset. The initial learning rate is 0.001, the batch size is 10 and the Adagrad \cite{duchi2011adaptive} optimizer is selected to optimize these two networks. For each time step, we input $12$ frames for the TSN network for the trajectory evaluation due to the limitation of our GPU. $\delta_1$ and $\delta_2$ are experimentally set as 0.5 and 0.6 in all the experiments. Our code is developed based on PyTorch and the experiments are conducted on a server with Ubuntu 16.04.3 LTS, Intel(R) Xeon(R) CPU E5-2620 v4@2.10GHz, GeForce RTX 2080. The source code is available at \textcolor{magenta}{\url{https://github.com/wangxiao5791509/DeepMTA_PyTorch}}.

\subsection{Comparison on Public Benchmarks}
In this section, we will report the tracking results of our method and other trackers on LaSOT, GOT-10k, UAV20L, OTB-2015, UAV123  and OxUvA datasets, respectively. It is worthy to note that we utilize our \textbf{THOR+BS-3T} version (termed DeepMTA) to compare with other trackers, although higher performance can be obtained with more trajectories.

\begin{figure}[!htb]
\center
\includegraphics[width=3.5in]{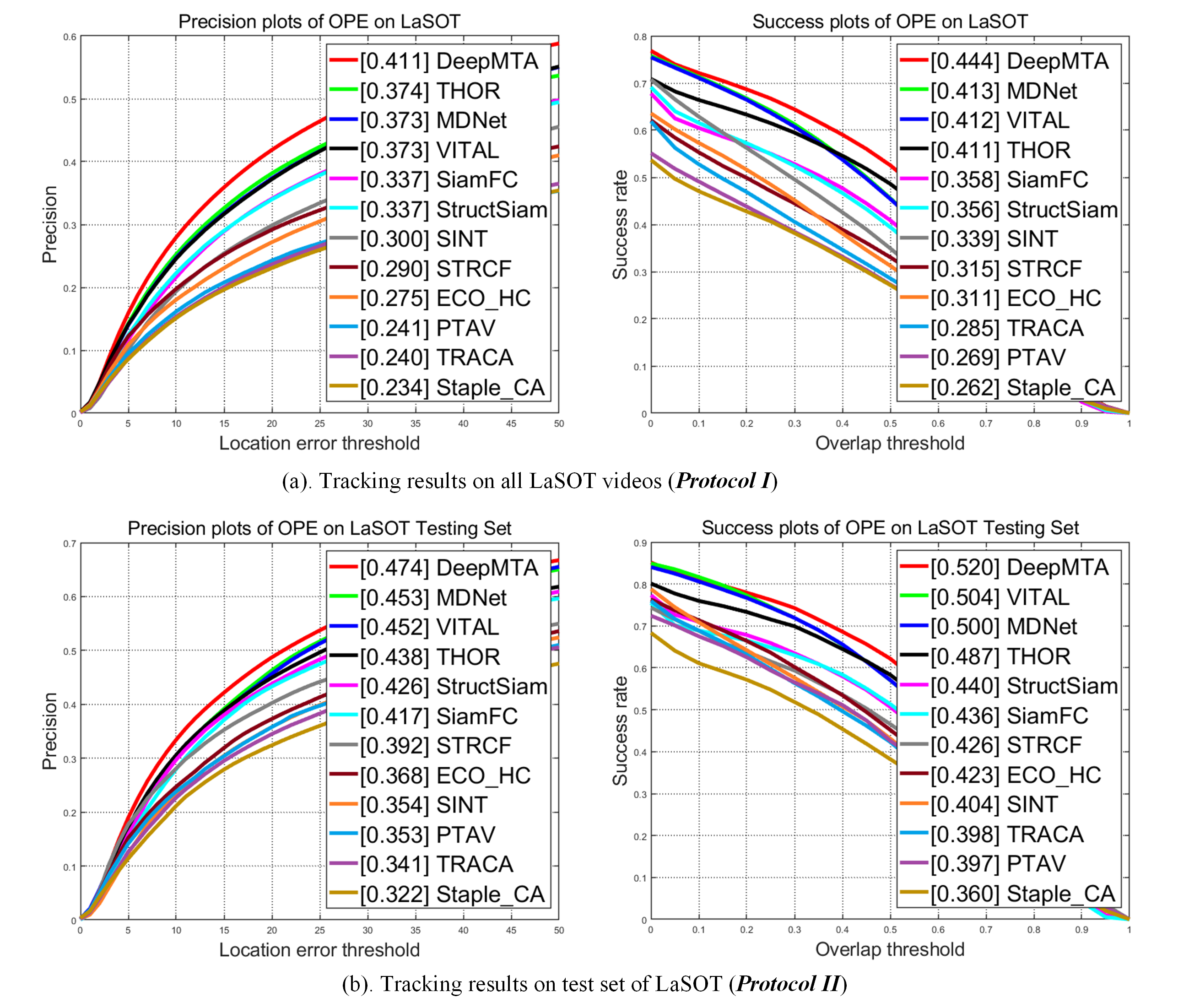}
\caption{ \textbf{Tracking results on LaSOT benchmarks. } DeepMTA is our proposed tracker in this paper. }
\label{lasot_results}
\end{figure}

\textbf{LaSOT \cite{fan2019lasot}:} As shown in Fig. \ref{lasot_results} (a), our tracker achieve 0.411 and 0.444 on precision plots and success plots based on the $Protocol I$, which are significantly better than the baseline THOR (0.374/0.411) and other trackers. For the $Protocol II$, our results are also better than these trackers as illustrated in Fig. \ref{lasot_results} (b).  This fully demonstrates the effectiveness of our proposed tracker.

\textbf{GOT-10k \cite{huang2018got}:} 
As we can see from Table \ref{GOT10kbenchmarkresults}, our tracker achieves better results than the baseline method THOR and also most of the other recent and popular trackers, such as MDNet and ECO \footnote{The related tracking results on GOT-10k are adopted from the leader board from \url{http://got-10k.aitestunion.com/leaderboard}. }. Specifically, the baseline tracker THOR achieve 0.447, 0.538 on the $AO$ and $SR_{0.50}$, respectively, while our tracker which utilize the dynamic target-aware attention guided multi-trajectory tracking obtained better results (0.462 and 0.556 on the evaluation metric). 

\begin{table*}[htp!]
\center
\scriptsize  
\caption{Tracking results on GOT-10k benchmark} \label{GOT10kbenchmarkresults}
\begin{tabular}{l|cccccccccccc}
\hline \toprule [1 pt]
\textbf{Tracker} 	&KCF	 \cite{Henriques2015High} &SRDCF \cite{danelljan2016adaptive}&DAT	 \cite{pu2018deep}&MDNet	 \cite{Nam2015Learning}	&ECO	 \cite{Danelljan2016ECO}	&GOTURN	 \cite{held2016GOTURN}&SiamFC \cite{bertinetto2016fully}	\\ 
\textbf{AO}      &0.203&0.236	&0.251  &0.299&0.316&0.347	&0.348	 \\ 
$\textbf{SR}_{0.50}$   &0.177	&0.227	&	0.242  &0.303	&	0.309	&0.375	 &0.353	 \\ 
\hline
\textbf{Tracker}			&ATOM \cite{danelljan2019atom}	&RT-MDNet \cite{Jung_2018_ECCV}  &THOR \cite{sauer2019tracking}&THOR+GS&THOR+BS-2T	&THOR+BS-3T	  &THOR+BS-3T-TSN  \\ 
\textbf{AO} 				&0.547			&0.342		 &0.447&0.453&0.458&0.461&0.462	\\ 
$\textbf{SR}_{0.50}$	&0.628			&0.356		 &0.538	   &0.545	 &0.550	   &0.554	 &0.556     \\ 		 	   
\hline \toprule [1 pt]
\end{tabular}
\end{table*}

\textbf{OTB-2015 \cite{wu2015object}: }   
As shown in Table \ref{OTBUAVresults}, THOR achieves 0.787/0.641 on the OTB-2015 dataset which are already better than most of the compared trackers, such as CFNet, SiamFC and Staple. Thanks to the dynamic target-aware attention, which can provide us global search regions for more robust tracking, we achieve 0.799/0.650 on this dataset. Although our overall performance is not better than some trackers (ECO: 0.910/0.691) on this dataset, however, our results are significantly better than these trackers on large-scale tracking benchmarks, such as LaSOT. We think this may be caused by overfitting of their tracker on this small and short-term dataset.

\begin{table*}[htp!]
\center
\scriptsize  
\caption{Tracking results on OTB-2015, UAV123 and UAV20L dataset.} \label{OTBUAVresults}
\begin{tabular}{c|ccccccc|cccccccc}
\hline \toprule [1 pt]
\textbf{OTB-2015}& {CFNet } &{SiamFC} &{ Staple} &{DSST } &{PTAV }  &{ECO}  &{CREST}  	&{THOR} 		&{Ours}  \\
\textbf{PR/SR} &0.748/0.568 &0.771/0.582  &0.784/0.581  &0.680/0.513 &0.848/0.634   &0.910/0.691    	&0.838/0.623   &0.787/0.641 &0.799/0.650 \\  
\hline \toprule [1 pt]
\textbf{UAV123} &{ DSST}     &{SAMF}   &{SRDCF}  &{ECO}   &{SiamRPN}             &{DaSiamRPN}      &{RT-MDNet}       &{THOR}        &{Ours}       \\  
\textbf{PR/SR} &0.586/0.356    &0.592/0.396      & 0.676/0.464     &0.741/0.525   &0.748/0.527     &0.796/0.586      &0.772/0.528      &0.758/0.697        &0.814/0.746    \\
\hline \toprule [1 pt]
\textbf{UAV20L} &{ PTAV}     &{SiamFC}  &{MUSTer}     &{SRDCF}   &{MEEM}  &{fDSST}    &{RT-MDNet}         &{THOR}     &{Ours}       \\
\textbf{PR/SR} &0.624/0.423     & 0.626/0.403     &0.514/0.329      & 0.507/0.343     &0.482/0.295     &0.422/0.300    &  0.583/0.461      &0.533/0.436       &0.715/0.570   \\
\hline \toprule [1 pt]
\end{tabular}
\end{table*}

\begin{table*}[htp!]
\center
\scriptsize 
\caption{Tracking results on OxUvA long-term benchmark.} \label{OxUvAresults}
\begin{tabular}{l|ccccccccccccc}
\hline \toprule [1 pt]
\textbf{Tracker} 	&SPLT \cite{yan2019skimming}&MBMD \cite{zhang2018learning}&SiamFC+R\cite{valmadre2018long}&TLD \cite{kalal2011tracking}&DaSiam LT\cite{zhu2018distractor} &LCT 	\cite{ma2015long}&LTSINT \cite{Tao2016Siamese}&MDNet 	\cite{Nam2015Learning} \\ 
\textbf{MaxGM}	&0.622  &0.544	&0.454  &0.431     &0.415    &0.396    &0.363     &0.343     \\ 
\textbf{TPR}      	&0.498	&0.609 	 &0.427      &0.208     &0.689   &0.292     &0.526    &0.472     \\
\hline
\textbf{Tracker} 	&SINT 	\cite{Tao2016Siamese}	&ECO-HC 	\cite{Danelljan2016ECO}&SiamFC 	\cite{bertinetto2016fully}	&EBT 	\cite{zhu2016beyond}&BACF 		\cite{kiani2017learning}&Staple 	\cite{bertinetto2016staple}	&THOR \cite{sauer2019tracking}&Ours  \\ 
\textbf{MaxGM}	&0.326    &0.314  &0.313    &0.283    &0.281    &0.261      &0.320   &0.340	 \\ 
\textbf{TPR}			&0.426     &0.395    &0.391     &0.321    &0.316   &0.273   &0.410  &0.463   \\ \hline \toprule [1 pt]
\end{tabular}
\end{table*}

\textbf{UAV123 \cite{mueller2016benchmarkuav20l}:} 
As shown in Table \ref{OTBUAVresults}, the baseline tracker THOR achieves 0.758/0.697 on PR and SR, which is comparable with existing trackers like RT-MDNet, SiamRPN and DaSiamRPN. Our tracker attains the best performance on this benchmark compared with these trackers, which are 0.814/0.746 on PR/SR, respectively. These experiments also fully demonstrate the effectiveness of our proposed dynamic target-aware attention guided multi-trajectory tracking framework.

\textbf{UAV20L \cite{mueller2016benchmarkuav20l}:} 
As shown in Table \ref{OTBUAVresults}, our tracker achieves 0.715/0.570 on the PR and SR, respectively, which are significantly better than the baseline tracker THOR and also some recent strong trackers like PTAV, RT-MDNet and SRDCF. This experiment fully validated the effectiveness of our proposed multi-trajectory tracking framework.

\textbf{OxUvA \cite{valmadre2018long}:} 
As shown in Table \ref{OxUvAresults}, we report the results on the test dataset of OxUvA which contains 166 long-term videos. The baseline method THOR achieves 0.320, 0.410 on MaxGM and TPR, while our tracker obtains 0.340 and 0.463.

\subsection{Ablation Study}

In this section, the following notations are needed to be watchful to better understand our model. Specifically, the MDC is short for the modules of convolutional operators between the feature maps of target template and global images. PP denotes the point proposal module, GS is short for global search mechanism, and BS means the beam search scheme for robust tracking, i.e., the multi-trajectory analysis module.
 
\textbf{Analysis on Dynamic Target-Aware Attention.} 
To check the effectiveness of each component of our target-aware attention model, we implement the following component analysis: 

1). ResNet+Concat: naive version for target-aware attention estimation. We directly concatenate the feature map of the target object and global image as previous work does \cite{wang2018describe}.    

2). ResNet+MDC: we conduct convolutional operations on multiple hierarchical feature maps, to check the effectiveness of interactions between target template and global images.  

3). ResNet+MDC+PP: we integrate the point proposal to check the influence of spatial coordinates. 

\begin{figure*}[!htb]
\center
\includegraphics[width=7in]{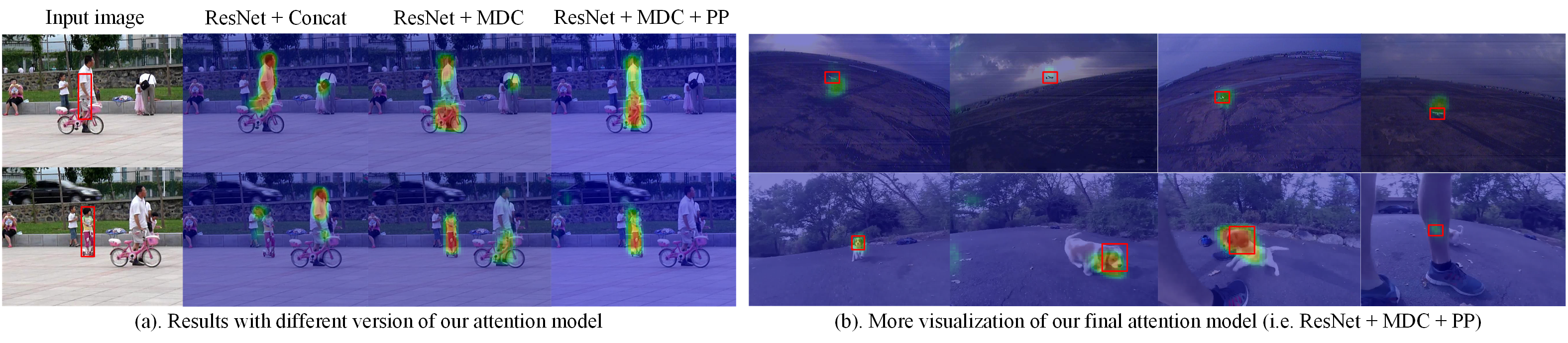}
\caption{ \textbf{Attention maps generated with different attention models}. }
\label{atten_visualization}
\end{figure*}

\begin{figure}[!htb]
\center
\includegraphics[width=3.5in]{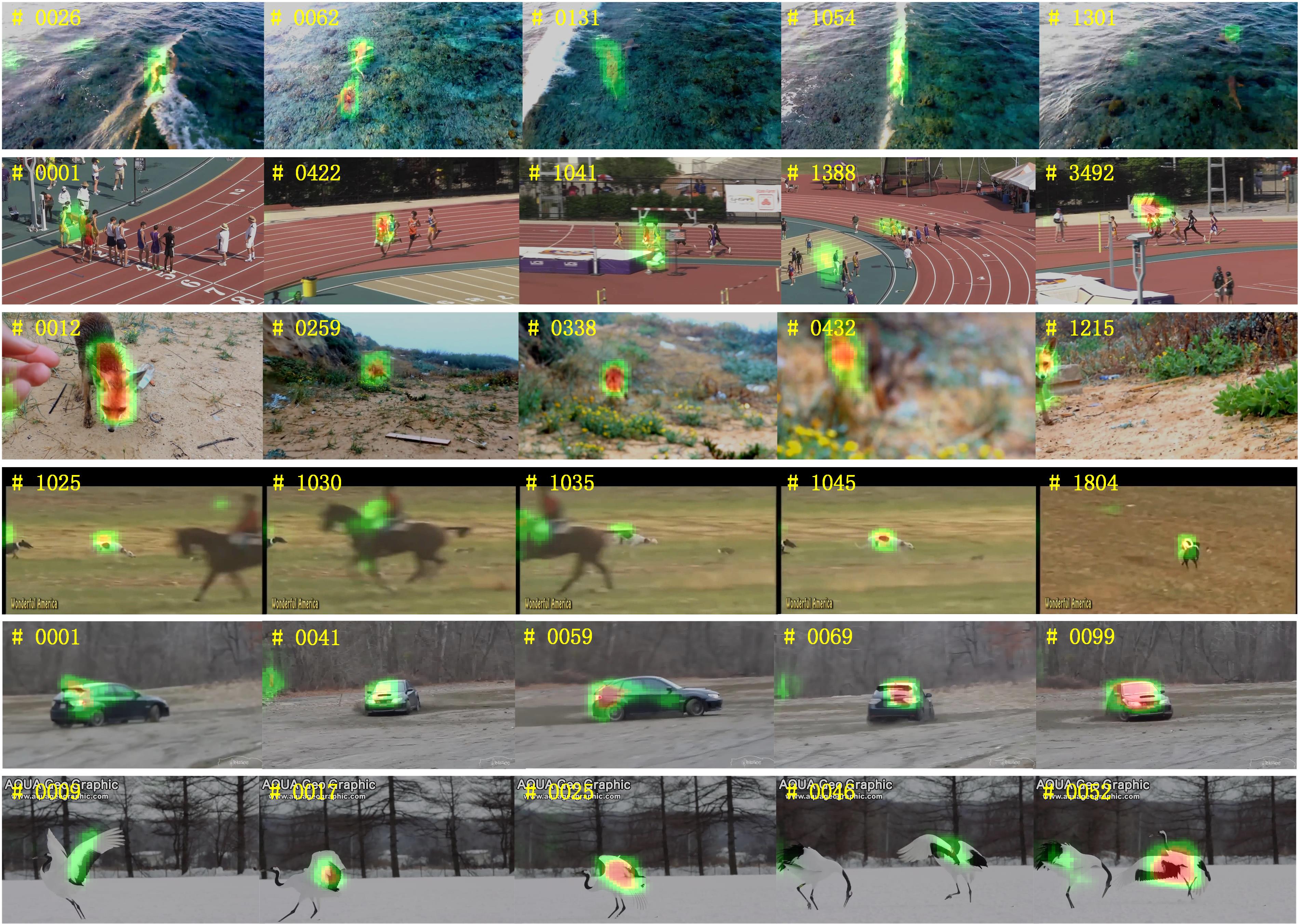}
\caption{ \textbf{Attention predicted by our dynamic target-aware attention network.} }
\label{attention_visualization}
\end{figure}

In this section, we utilize MAE (Mean Absolute Error), which is a widely used evaluation metric in the salient object detection community \cite{chen2020recursive, chen2019shape}, to measure the quality of predicted attention maps of each model. As shown in Table \ref{MAEResults}, the MAE of ResNet+Concat is 4.27, while the ResNet+MDC and ResNet+MDC+PP are 3.99 and 2.23, respectively. It is easy to find that the introduced convolutional operation and spatial coordinates are all contributed to the dynamic target-aware attention estimation. Some attention maps predicted with these models can be found in Fig. \ref{atten_visualization}, and more results are provided in Fig. \ref{attention_visualization}.

\begin{table}[htp!]
\center
\scriptsize 
\caption{MAE of attention prediction on OTB-2015 dataset.} \label{MAEResults}
\begin{tabular}{c|c|c|c}
\hline \toprule [1 pt]
Model    &ResNet+Concat  &ResNet+MDC &ResNet+MDC+PP  \\ 
\hline
MAE   	&4.27    &3.99   &2.23        \\
\hline \toprule [1 pt]
\end{tabular}
\end{table}

\textbf{Analysis on Multi-Trajectory Tracking.} 
To check the effectiveness of our proposed multi-trajectory inference strategy, we conduct the following component analysis: 

i). THOR: the baseline method used in this paper; 

ii). THOR+GS: we introduce target-aware attention for global search and integrate with THOR to check the effectiveness of target-aware attention; 

iii). THOR+BS-2T/3T: we utilize \emph{double/triplet-trajectory} search strategy for component-ii to check the influence of multi-trajectory tracking; 

iv). THOR+BS-3T-TSN: we use the trajectory selection network (TSN) for final trajectory selection to check the effectiveness of this module.

As shown in Table \ref{GOT10kbenchmarkresults}, the baseline tracker THOR achieves 0.447/0.538 on AR and SR, respectively. When integrating our dynamic target-aware attention module into THOR, the tracking performance improved to 0.453/0.545, which fully demonstrates the effectiveness of dynamic target-aware attention for global search. We also conduct tracking based on our proposed multi-trajectory tracking framework and further improve the tracking results. For example, the THOR+BS-2T/3T improves the tracking results from 0.453/0.545 to 0.458/0.550 and 0.461/0.554, respectively. These experiments fully validated the effectiveness of our multi-trajectory tracking framework. It is also worthy to note that the aforementioned trackers are all select the final trajectory based on their response score only. However, the response score can not reflect the true results in some complex scenarios. The performance can be further improved by using our proposed trajectory selection network, which not only considers the response score from tracker but also the target-aware attention and consistency of the tracked target object. The results of THOR+BS-3T-TSN  fully validated the effectiveness of this module.

From a methodological point of view, more trajectories mean more dynamics we can capture with our model. As shown in Table \ref{GOT10kbenchmarkresults}, we can find that the tracking results can be further improved by introducing more trajectories. Specifically, the tracking results of 1, 2 and 3 trajectories on the GOT-10k dataset are 0.453/0.545, 0.458/0.550 and 0.461/0.554 which are consistent with our views. In addition, we also evaluate different trajectories (3, 4, 5 trajectories are tested) based on OTB-2015 dataset and we get 0.799/0.650, 0.799/0.650, 0.799/0.651, respectively. These experiments all demonstrate that beam search based multi-trajectory tracking achieves better results than regular greedy search strategy.

\textbf{Analysis on the Generalization.} 
In our experiments, we select the THOR as our baseline tracker for tracking. It is worthy to note that our proposed algorithm can also be integrated with other tracking algorithms due to it is a generic module. In this section, we test our module by integrating them with SiamFC++ \cite{xu2020siamfc++}, SiamMask \cite{wang2019fast}, DiMP \cite{bhat2019DiMP}, MDNet \cite{Nam2015Learning}, SiamRPN++ \cite{li2018siamrpn++}, and ATOM \cite{danelljan2019atom} on the GOT-10k and VOT2018LT dataset to check the generalization.

As shown in Table \ref{Generalization}, we can find that our module can improve all the baseline approaches on both datasets. Specifically, the SiamFC++ achieves 0.604/0.734 on the AO/SR respectively, while our method attain 0.609/0.738 on the GOT-10k dataset. We also improve the SiamMask from 0.451/0.541 to 0.461/0.555, the MDNet from 0.299/0.303 to 0.392/0.433. The improvement of SiamFC++ and SiamMask on the GOT-10k dataset is relatively little, due to the videos are short, therefore, there is little room for our module to improve final results. For the long-term benchmark VOT2018LT which contains 35 videos, we can find that the improvements are significant. More detail, the SiamFC++ achieves 0.689, 0.457, 0.549 on precision/recall/F1-score respectively, meanwhile, we improve these metrics to 0.684, 0.544, 0.606. For the minor decrease of SiamFC++ on the Precision, we think this maybe caused by the tradeoff between local search and global search mechanism. The experimental results based on MDNet, SiamMask, DiMP, ATOM, and SiamRPN++ trackers also demonstrate that our proposed modules are effective for tracking task, especially on the long-term video sequences.

\begin{table*}[htp!]
\center
\scriptsize
\caption{Generalization analysis on the GOT-10k and VOT2018-LT dataset.} \label{Generalization}
\begin{tabular}{c|cc|cc|cc|cc|cc|cc} 
\hline \toprule [1 pt]
\textbf{GOT-10k}   	&\textbf{SiamFC++} \cite{xu2020siamfc++}     &\textbf{Ours}           &\textbf{MDNet} \cite{Nam2015Learning}    &\textbf{Ours}           &\textbf{SiamMask} \cite{wang2019siamMask}    &\textbf{Ours}      &\textbf{DiMP}	\cite{bhat2019DiMP}		&\textbf{Ours}        &\textbf{ATOM} \cite{danelljan2019atom}			&\textbf{Ours}     &\textbf{SiamRPN++} \cite{li2018siamrpn++}			&\textbf{Ours}          \\ 
\hline 
\textbf{AO}   	    	&0.604      		&0.609          &0.299     &0.392          		&0.451        &0.461	        &0.673				&0.674				&0.547				&0.557				&0.453				&0.473						   \\
\textbf{SR}         		&0.734      		&0.738          &0.303     &0.433          		&0.541        &0.555         &0.785				&0.788				&0.628				&0.645				&0.537				&0.561				   \\
\hline \toprule [1 pt]
\textbf{VOT2018LT}      	&\textbf{SiamFC++} \cite{xu2020siamfc++}     &\textbf{Ours}           &\textbf{MDNet} \cite{Nam2015Learning}    &\textbf{Ours}           &\textbf{SiamMask} \cite{wang2019siamMask}    &\textbf{Ours}      &\textbf{DiMP}	\cite{bhat2019DiMP}		&\textbf{Ours}        &\textbf{ATOM} \cite{danelljan2019atom}			&\textbf{Ours}     &\textbf{SiamRPN++} \cite{li2018siamrpn++}			&\textbf{Ours}          \\ 
\hline 
\textbf{Precision}      		&0.689      &0.684          &0.479         &0.562         &0.610         &0.615           &0.660				&0.670				&0.619				&0.625				&0.646				&0.659				   \\
\textbf{Recall}        	 		&0.457      &0.544          &0.324         &0.355          &0.407         &0.498          &0.583				&0.585				&0.485				&0.507				&0.419				&0.442				   \\
\textbf{F1-score}       		&0.549      &0.606          &0.387         &0.435          &0.488         &0.550          &0.619				&0.625				&0.544				&0.560				&0.508				&0.529				   \\
\hline \toprule [1 pt]
\end{tabular}
\end{table*}

\textbf{Efficiency Analysis.} 
The baseline tracker THOR running at 112 FPS reported in their original paper (named as THOR-I in Table \ref{fpsResults}); while it running at 79.73 FPS on a Laptop with CPU Intel I7 and GPU NVIDIA RTX 2070 (i.e., the THOR-II in Table \ref{fpsResults}). This speed is tested on the whole OTB-2015 dataset. Our tracker can run at 12.10 and 12.07 FPS respectively when 2 and 3 trajectories are adopted. It is also worthy to note that this running time includes both the \emph{multi-trajectory tracking} and \emph{multi-trajectory selection}. We believe that our tracker can obtain better running efficiency, if better GPU is utilized, such as NVIDIA RTX 2080. 
 
\begin{table}[htp!]
\center
\scriptsize 
\caption{Efficiency Analysis of our model.} \label{fpsResults}
\begin{tabular}{c|ccccccc}
\hline \toprule [1 pt]
\textbf{Tracker}    &\textbf{THOR-I}  &\textbf{THOR-II} &\textbf{Ours-2}  &\textbf{Ours-3} \\ 
\hline 
\textbf{FPS}   	 	&112	 &79.73			&12.10			&12.07    \\
\hline \toprule [1 pt]
\end{tabular}
\end{table}

\textbf{Influence of Threshold Parameters. } 
In this work, two parameters $\delta_1$ and $\delta_2$ are very important for final tracking. We report the tracking results on OTB-2015 dataset with different settings in Table \ref{thresholdAnalysisResults}. Specifically speaking, we first fix $\delta_2$ as 0.6, and test the performance with various values of $\delta_1$, i.e., from 0.3 to 0.9. We can find that the results will be better when the $\delta_1$ is 0.4. Then, we fix the $\delta_1$ as 0.4, and check the results with various values of $\delta_2$. Finally, it is easy to find that we can attain the best performance when $\delta_1$ and  $\delta_2$ are set as 0.4 and 0.6. Meanwhile, we can also find that our tracker is not so sensitive to these two parameters. Because the results are relatively stable when the value changing from 0.3 to 0.7. 
\begin{table}[htp!]
\center
\scriptsize 
\caption{Results with different threshold parameters $\delta_1$ and $\delta_2$ on OTB-2015 dataset.} \label{thresholdAnalysisResults}	 
\begin{tabular}{c|cccccccc}
\hline \toprule [1 pt]
$\delta_1$    &Baseline		&0.3				&0.4			&0.5			&0.6			&0.7			&0.8			&0.9			\\ 
\hline
PR   	 			&0.787			&0.799			&0.799		&0.796		&0.795		&0.795		&0.786		&0.769			\\
SR   	 			&0.641			&0.650			&0.650		&0.649		&0.648		&0.647		&0.642		&0.629			\\
\hline \toprule [1 pt]
$\delta_2$    &Baseline			&0.3			&0.4			&0.5			&0.6			&0.7			&0.8			&0.9			\\ 
\hline
PR   	 			&0.787				&0.791				&0.794				&0.796				&0.799				&0.798				&0.795				&0.788			\\
SR   	 			&0.641				&0.645				&0.646				&0.649				&0.650				&0.649				&0.648				&0.643			\\
\hline \toprule [1 pt]
\end{tabular}
\end{table}

\begin{figure}[!htb]
\center
\includegraphics[width=3.3in]{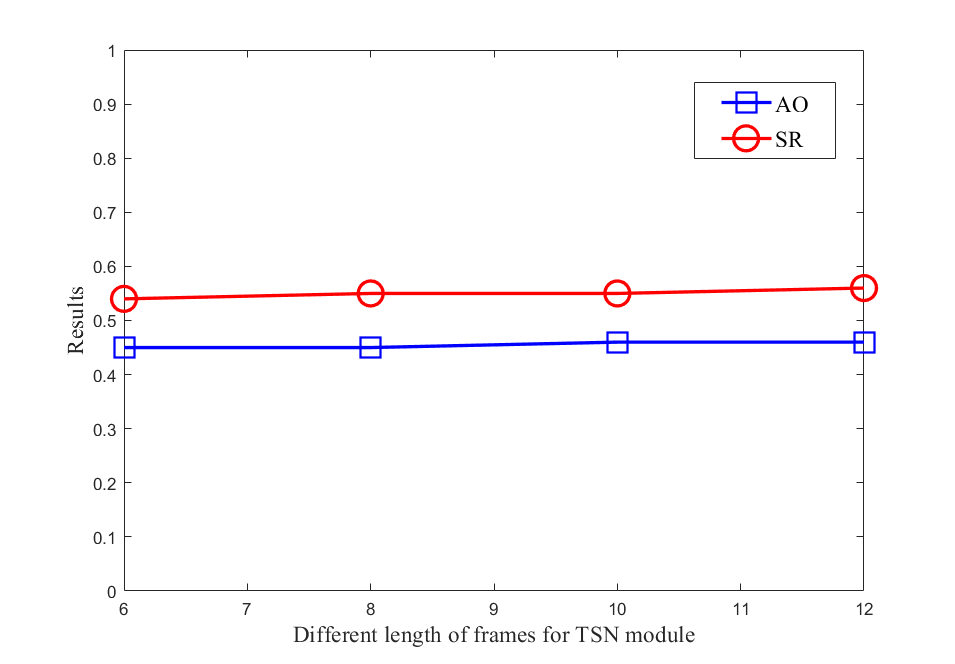}
\caption{ \textbf{Results of different number of frames for TSN module on GOT-10k dataset. }}
\label{param1}
\end{figure}

\textbf{Analysis on Different Number of Frames for TSN Module. } 
In our experiments, the TSN network is used for trajectory evaluation by processing the whole trajectory in a batch manner, i.e., we set the batch size as 12, due to the limited memory of our GPU. Much larger frame number can also be used, if the GPU meets the demand. To analyse the influence of this parameter, we also tested other values on GOT-10k dataset. From Fig. \ref{param1}, we find that the overall results are relatively stable when different values are used, i.e., 6, 8, 10, 12. 

\begin{figure*}[!htb]
\center
\includegraphics[width=7in]{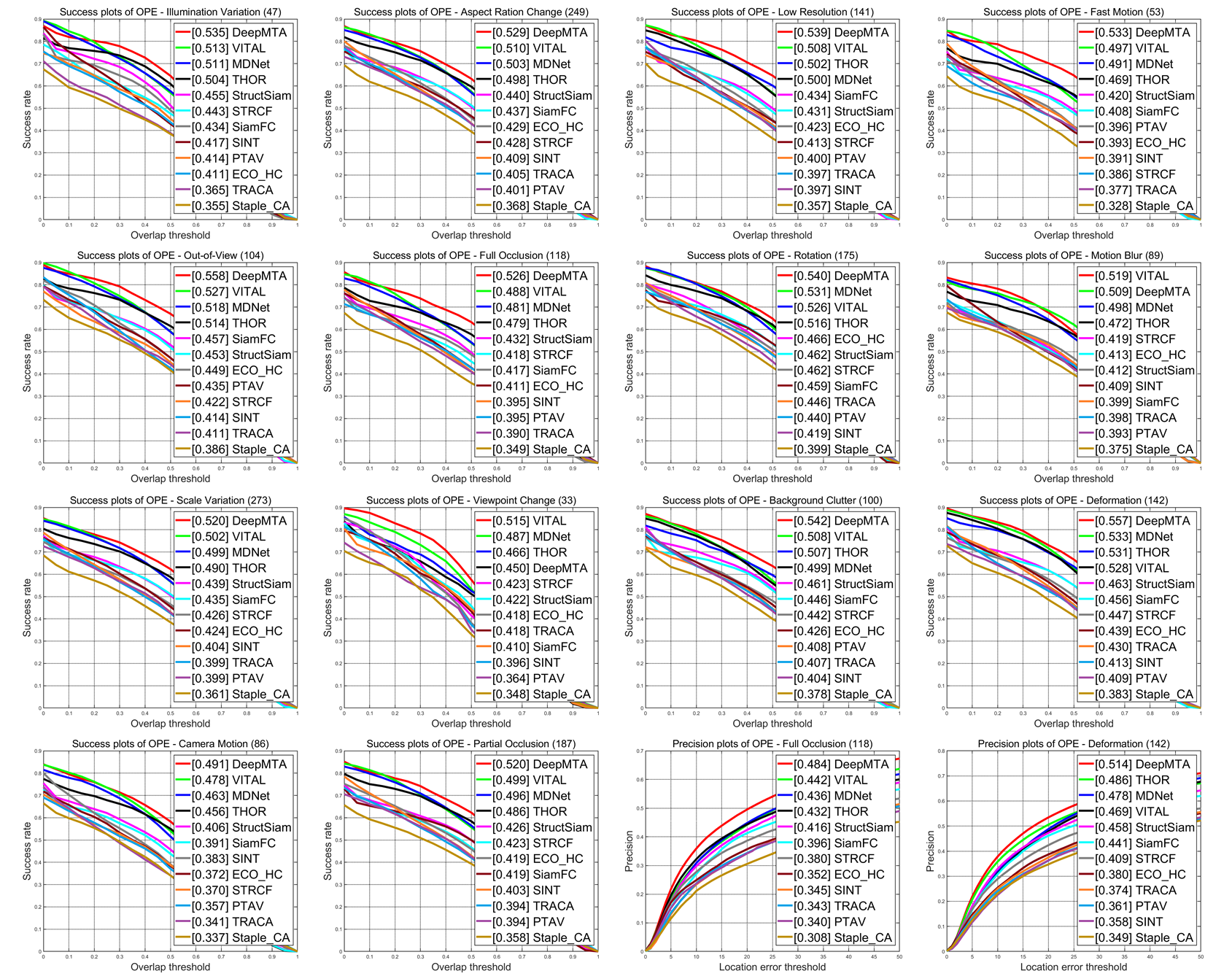}
\caption{  \textbf{Success plots of attribute analysis on LaSOT benchmarks. } Two sub-figures from Precision plots are moved in this figure for aesthetics.}
\label{lasot_attributes_1}
\end{figure*}

\begin{figure*}[!htb]
\center
\includegraphics[width=7in]{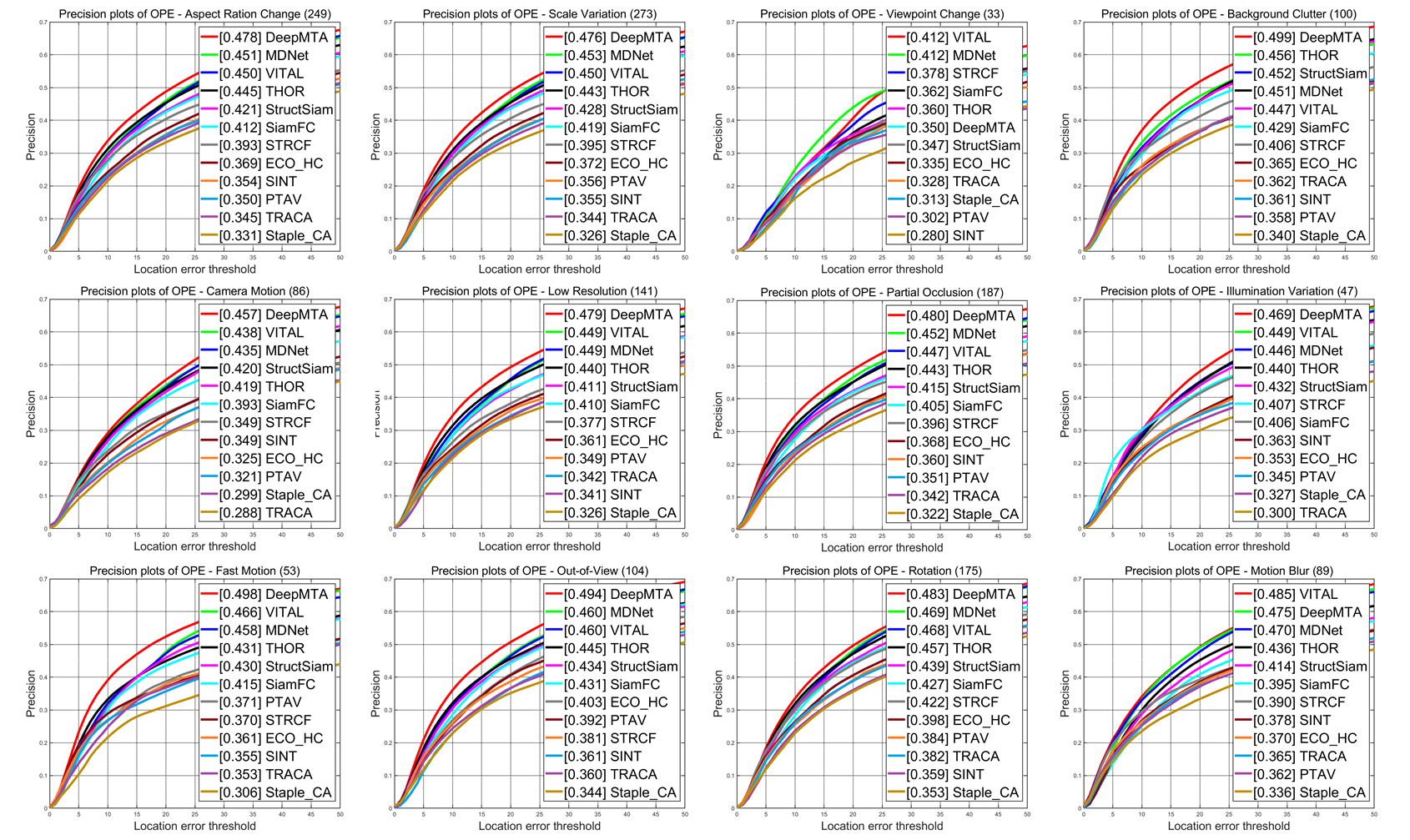}
\caption{ \textbf{ Precision plots of attribute analysis on LaSOT benchmarks.} }
\label{lasot_attributes_2}
\end{figure*}

\textbf{Attribute Analysis.} \label{attributeAnalysis}
In this section, we report the results of our tracker and some state-of-the-art trackers on each attribute as shown in Fig. \ref{lasot_attributes_1} (Success plots) and Fig. \ref{lasot_attributes_2} (Precision plots). It is easy to find that our tracker achieves the best performance on most of attributes, including out-of-view, low resolution, aspect ratio change. These experiments fully validated the effectiveness and robustness of our tracker when facing challenging factors.

\textbf{Visualization.}  
In this section, we give some visualization of our attention maps and tracking results respectively. 

\emph{Attention Maps:}
As we can see from Fig. \ref{attention_visualization}, our dynamic target-aware attention network can locate the attention regions which most related to the target object we want to track for each video. Our model show good robustness to \emph{clutter background, heavy occlusion, motion blur} and \emph{view rotation}, \emph{etc}. Specifically speaking, our attention model can still locate the target object accurately in the clutter background, such as the \emph{shark} and \emph{person} in the first and second row, respectively. This fully validated the effectiveness of our convolutional operation based target-aware attention prediction. It is also worthy to note that our attention can reflect the occluded target object to some extent, such as the \emph{shark} in the $0026$ and $1054$ frame at the first row, the \emph{dog} in the $1030$ at the fourth row. These amazing attention results are brought by our dynamic point proposal, which can provide spatial coordinate information for a more accurate target object location. Besides, the \emph{fox} and \emph{car} in the third and fifth row demonstrate that our dynamic attention model is robust to the view variation of the target object. 

According to the aforementioned analysis, we can conclude that our dynamic attention model shows good robustness to challenging factors, such as motion blur, heavy occlusion, out-of-view, clutter background. The tracking results on each attribute on the LaSOT dataset also proved the robustness of our model, as shown in Section \ref{attributeAnalysis}.

\begin{figure}[!htb]
\center
\includegraphics[width=3.5in]{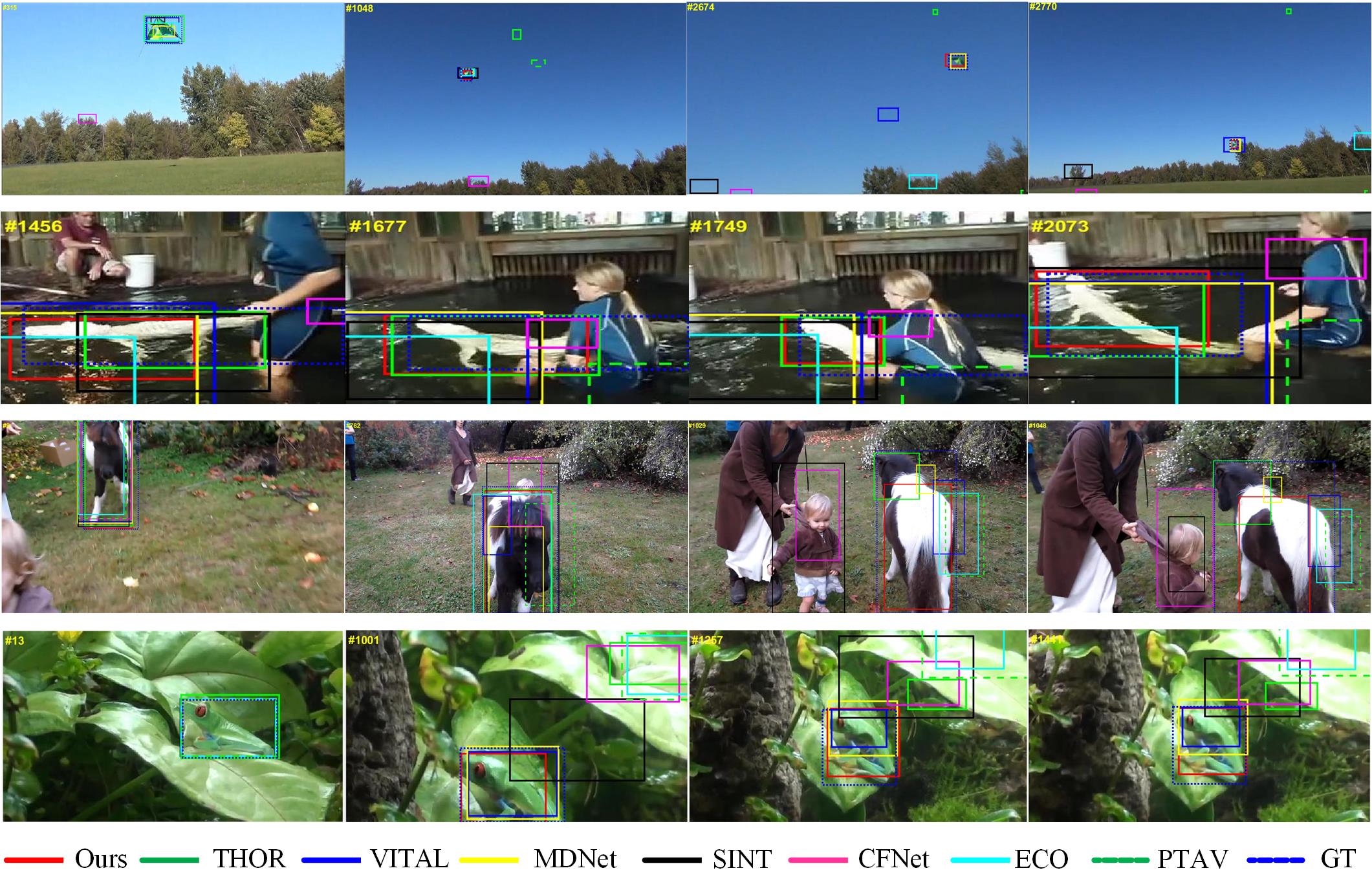}
\caption{ \textbf{Visualization of tracking results on videos from LaSOT benchmarks.} }
\label{lasot_visualization}
\end{figure}

\emph{Tracking Results:}
As shown in Fig. \ref{lasot_visualization}, we give some visualizations of our tracker and other state-of-the-art trackers on the LaSOT dataset. It is intuitive to find that our tracker is robust to challenging factors such as \emph{out-of-view, clutter background, scale variation}. For example, the \emph{flying kite} in the first row will become out of the view, when it occurred back, our tracker can still capture its location due to the utilize of joint local and global search scheme. However, many other trackers failed to locate the target object due to only the local search mechanism used in their procedure. This also demonstrates the importance of accurate prediction of dynamic target-aware attention maps. 

For the second and third row, we can find that our tracker (red bounding box) can locate the target object more accurately than the baseline tracker THOR (green bounding box). This fully validates the effectiveness of our tracking algorithm. For the fourth row, we can find that our tracker can still work well in challenging scenarios, while many other trackers (including the baseline tracker) are easily influenced by the clutter background. 
 
\subsection{Discussion} \label{Discussion}
The baseline tracker THOR only use local search under tracking-by-detection framework; the target-aware attention model provides the global attention map which can be used for global search for baseline tracker. On the other hand, the baseline tracker can provides tracking results to dynamically modeling the target object in the target-aware attention module. Therefore, these two modules are complementary to each other. Tracking by switching between local and global search is an intuitive way for robust object tracking, however, this may still confuse trackers when challenging factors occurred. Therefore, the multi-trajectory analysis module is introduced to collect multi-trajectories and conduct tracking by multi-trajectory selection. This procedure is mainly implemented by the trajectory selection network. It is also worthy to note that our algorithm tracking the target object in an \emph{batch} manner which will be beneficial for specific applications, such as video analysis in sport and surveillance.

\section{Conclusion and Future Work} 
In this paper, we propose a novel multi-trajectory tracking framework, which significantly increased the dynamics of visual tracking. Specifically, we maintain multiple tracking results for each frame based on joint local and global search. To conduct a more accurate global search, we design a novel dynamic target-aware attention module which receives dynamic target templates and coordinates as condition and estimate target location from global views. After all the video frames are processed, we select the best trajectory with our proposed trajectory selection network, which considers multiple information, such as attention maps, response scores, and coordinates of BBox. Extensive experiments are conducted on multiple tracking dataset, including short-term and long-term tracking datasets.

In our implementation, we simply select the best-scored trajectory as our tracking result, but different trajectories may have their own good tracking result clips. How to design an efficient and effective trajectory fusion scheme to achieve better tracking performance is a worthy study problem. The efficiency of our tracker can also be improved by adaptively choosing the number of trajectories. In other words, limited trajectories are needed for simple videos and more trajectories can be employed for challenging videos. We will focus on these two issues in our future works.

\section*{Acknowledgements}	
This work is jointly supported by Key-Area Research and Development Program of Guangdong Province 2019B010155002, Postdoctoral Innovative Talent Support Program BX20200174, China Postdoctoral Science Foundation Funded Project 2020M682828, Australian Research Council Projects FL-170100117, National Nature Science Foundation of China (61860206004, 61825101, 62076003).

{
\bibliographystyle{IEEEtran}
\bibliography{reference}
}

% that's all folks
\end{document}